\documentclass{article}

\usepackage[margin=1in]{geometry}

\usepackage{graphicx}
\usepackage{booktabs}
\usepackage{amsmath}
\usepackage{amssymb}
\usepackage{amsfonts}
\usepackage{multirow}
\usepackage{verbatim}
\usepackage{caption}
\usepackage{longtable}
\usepackage{supertabular}
\usepackage{wrapfig}
\usepackage{float}

\usepackage{enumitem}
\usepackage{tablefootnote}
\usepackage[round,semicolon]{natbib}
\usepackage{xcolor}
\usepackage{xspace}
\usepackage{textcomp}
\usepackage{makecell}
\usepackage{multirow}
\usepackage{lscape} 
\usepackage{siunitx}

\usepackage{amssymb}
\usepackage{pifont}
\usepackage{scrextend}

\usepackage{array}
\usepackage{tgpagella}
\usepackage{latexsym}
\usepackage[T1]{fontenc}
\usepackage[utf8]{inputenc}
\usepackage{microtype}
\definecolor{mydarkblue}{rgb}{0,0.08,0.45}
\usepackage[colorlinks,citecolor=mydarkblue,urlcolor=mydarkblue,linkcolor=mydarkblue]{hyperref}
\usepackage{url}            
\usepackage{nicefrac}       
\usepackage{changepage}
\usepackage{xargs}          
\usepackage{wrapfig,lipsum,booktabs}
\usepackage{longtable}
\usepackage{subcaption}
\usepackage{endnotes}

\usepackage{pgfplots}
\usetikzlibrary{pgfplots.groupplots}
\pgfplotsset{compat=1.3}
\usepackage{tikz}
\usetikzlibrary{patterns}

\usepackage[most]{tcolorbox}

\usepackage[capitalize,noabbrev]{cleveref}
\crefname{section}{Section}{\S\S}
\Crefname{section}{Section}{\S\S}
\crefname{table}{Table}{Tables}
\crefname{figure}{Figure}{Figures}
\crefname{algorithm}{Algorithm}{}
\crefname{equation}{eq.}{}
\crefname{appendix}{Appendix}{}
\crefformat{section}{Section #2#1#3}
\usepackage{multicol}

\definecolor{battleshipgrey}{rgb}{0.3, 0.3, 0.3}
\definecolor{brilliantrose}{rgb}{1.0, 0.33, 0.64}
\definecolor{americanrose}{rgb}{1.0, 0.01, 0.24}
\definecolor{jweigreen}{rgb}{0,0.45,0.24}
\definecolor{bluegray}{rgb}{0.1, 0.1, 0.4}
\definecolor{ao(english)}{rgb}{0.0, 0.5, 0.0}
\definecolor{blanchedalmond}{rgb}{1.0, 0.92, 0.8}
\definecolor{atomictangerine}{rgb}{1.0, 0.6, 0.4}
\definecolor{chocolate(web)}{rgb}{0.82, 0.41, 0.12}
\definecolor{bananayellow}{rgb}{1.0, 0.88, 0.21}
\definecolor{goldenbrown}{rgb}{0.6, 0.4, 0.08}
\definecolor{aliceblue}{rgb}{0.94, 0.97, 1.0}
\definecolor{beige}{rgb}{0.96, 0.96, 0.86}
\definecolor{babyblue}{rgb}{0.54, 0.81, 0.94}
\definecolor{camel}{rgb}{0.76, 0.6, 0.42}
\definecolor{cinnamon}{rgb}{0.82, 0.41, 0.12}
\definecolor{deepskyblue}{rgb}{0.0, 0.75, 1.0}
\definecolor{frenchblue}{rgb}{0.0, 0.45, 0.73}
\definecolor{classicrose}{rgb}{0.98, 0.8, 0.91}
\definecolor{frenchrose}{rgb}{0.96, 0.29, 0.54}
\definecolor{frenchlilac}{rgb}{0.53, 0.38, 0.56}
\definecolor{frenchbeige}{rgb}{0.65, 0.48, 0.36}

\definecolor{forestgreen}{HTML}{2e7d43}
\definecolor{color1}{HTML}{FF9999}
\definecolor{color2}{HTML}{FF6666}
\definecolor{color3}{HTML}{FF3333}
\definecolor{color4}{HTML}{E60000}
\definecolor{color5}{HTML}{B30000}
\definecolor{color6}{HTML}{8CD98C}
\definecolor{color7}{HTML}{53c653}
\definecolor{color8}{HTML}{39ac39}
\definecolor{color9}{HTML}{2d862d}
\definecolor{color10}{HTML}{206020}
\definecolor{color11}{HTML}{cca300}

\newcommand{\palm}[0]{PaLM}

\newcommand{\tzeromixture}[0]{T0-SF}

\newcommand{\flantwo}[0]{Flan 2022}

\usepackage{minitoc}

\title{
\vspace{-10mm}
\textbf{
The Flan Collection: Designing Data and Methods \\for Effective Instruction Tuning
}
\vspace{-3mm}
  }
  
\author{
\normalsize{}
\textbf{Shayne Longpre\thanks{Research completed while a Student Researcher at Google. Correspondence: \url{slongpre@mit.edu}.}} \hspace{5mm}
\textbf{Le Hou} \hspace{5mm} 
\textbf{Tu Vu} \hspace{5mm}
\textbf{Albert Webson} \hspace{5mm}
\textbf{Hyung Won Chung} \hspace{5mm} 
\\
\normalsize{}
\textbf{Yi Tay} \hspace{4mm}
\textbf{Denny Zhou} \hspace{4mm} 
\textbf{Quoc V. Le} \hspace{4mm}
\textbf{Barret Zoph} \hspace{5mm}
\textbf{Jason Wei} \hspace{5mm} 
\textbf{Adam Roberts} \hspace{4mm}
\\
\\
\normalsize{}
Google Research
\vspace{-4mm}
}

\date{}

\begin{document}

\doparttoc 
\faketableofcontents 

\maketitle

\begin{abstract}
\noindent 

We study the design decisions of publicly available instruction tuning methods, and break down the development of \flantwo{} models \citep{chung2022scaling}.
Through careful ablation studies on the Flan Collection \emph{of instruction tuning tasks and methods}, we tease apart the effect of design decisions that enable Flan-T5 to outperform prior work by 3-17\%+ across evaluation settings. 
We find task balancing and enrichment techniques are overlooked but critical to effective instruction tuning, and in particular, training with mixed prompt settings (zero-shot, few-shot, and chain-of-thought) actually yields stronger (2\%+) performance in \emph{all} settings.
In further experiments, we show Flan-T5 requires less finetuning to converge higher and faster than T5 on single downstream tasks---motivating instruction-tuned models as more computationally-efficient starting checkpoints for new tasks.
Finally, to accelerate research on instruction tuning, we make the \flantwo{} collection of datasets, templates, and methods publicly available.\footnote{Data generation code available at: \url{https://github.com/google-research/FLAN/tree/main/flan/v2}. Generation code allows users to vary mixtures rates, templates, prompt types and data augmentations techniques, for faster public research.}

\end{abstract}

\makeatletter
\newenvironment{customlegend}[1][]{%
    \begingroup
    \pgfplots@init@cleared@structures
    \pgfplotsset{#1}%
}{%
    \pgfplots@createlegend
    \endgroup
}%

\def\addlegendimage{\csname pgfplots@addlegendimage\endcsname}

\begin{figure}[ht]
\pgfplotsset{width=4.3cm, height=6.0cm,
    /pgfplots/ybar legend/.style={
    /pgfplots/legend image code/.code={%
       \draw[##1,/tikz/.cd,yshift=-0.25em]
        (0cm,0cm) rectangle (7pt,0.8em);},
   },}
    \centering
    \begin{tikzpicture}  
    \begin{groupplot}[
          group style={
          group name=plot,
          horizontal sep=6pt,
          vertical sep=0pt,
          group size=5 by 1},]
      \nextgroupplot[
            ybar,
            ymin=0, ymax=80,
            ytick={0, 20, 40, 60, 80},
            major x tick style = transparent,
            bar width=12pt,
            enlarge x limits=0.25,
            ylabel={Avg. Accuracy (\%)},
            symbolic x coords={Zero-Shot Held-In},  
            xtick=data,  
            axis x line*=bottom,
            axis y line*=left,
            y label style={at={(axis description cs:-0.22,0.5)},anchor=south},
            ]  
        \addplot[ybar, fill=frenchblue,  postaction={}] coordinates {
            (Zero-Shot Held-In, 73.8)
        };
        \addplot[ybar, fill=babyblue,  postaction={}] coordinates {
            (Zero-Shot Held-In, 68.4)
        };  
        \addplot[ybar, fill=classicrose,  postaction={}] coordinates {
            (Zero-Shot Held-In, 70.6)
        };  
        \addplot[ybar, fill=beige,  postaction={}] coordinates {
            (Zero-Shot Held-In, 50.3)
        };  
 
      \nextgroupplot[
            ybar,
            ymin=0, ymax=80,
            ytick={0, 10, 20, 30, 40, 50, 60, 70, 80},
            major x tick style = transparent,
            bar width=12pt,
            enlarge x limits=0.25,
            symbolic x coords={Zero-Shot CoT},  
            xtick=data,  
            axis x line*=bottom,
            axis y line*=left,
            y axis line style={opacity=0},
            ytick=\empty,
            ]  
        \addplot[ybar, fill=frenchblue,  postaction={}] coordinates {
            (Zero-Shot CoT, 34.2)
        };
        \addplot[ybar, fill=babyblue,  postaction={}] coordinates {
            (Zero-Shot CoT, 24.6)
        };  
        \addplot[ybar, fill=classicrose,  postaction={}] coordinates {
            (Zero-Shot CoT, 25.6)
        };  
        \addplot[ybar, fill=beige,  postaction={}] coordinates {
            (Zero-Shot CoT, 13.8)
        };  
        
      \nextgroupplot[
            ybar,
            ymin=0, ymax=80,
            ytick={0, 10, 20, 30, 40, 50, 60, 70, 80},
            major x tick style = transparent,
            bar width=12pt,
            enlarge x limits=0.25,
            symbolic x coords={Few-Shot BBH},  
            xtick=data,  
            axis x line*=bottom,
            axis y line*=left,
            y axis line style={opacity=0},
            ytick=\empty,
            ]  
        \addplot[ybar, fill=frenchblue,  postaction={}] coordinates {
            (Few-Shot BBH, 39.3)
        };
        \addplot[ybar, fill=babyblue,  postaction={}] coordinates {
            (Few-Shot BBH, 28.3)
        };  
        \addplot[ybar, fill=classicrose,  postaction={}] coordinates {
            (Few-Shot BBH, 30.8)
        };  
        \addplot[ybar, fill=beige,  postaction={}] coordinates {
            (Few-Shot BBH, 15.6)
        };  
        \addplot[ybar, fill=gray!35,  postaction={}] coordinates {
            (Few-Shot BBH, 35.7)
        };  

      \nextgroupplot[
            ybar,
            ymin=0, ymax=80,
            ytick={0, 10, 20, 30, 40, 50, 60, 70, 80},
            major x tick style = transparent,
            bar width=12pt,
            enlarge x limits=0.25,
            symbolic x coords={Zero-Shot MMLU},  
            xtick=data,  
            axis x line*=bottom,
            axis y line*=left,
            y axis line style={opacity=0},
            ytick=\empty,
            ]  
        \addplot[ybar, fill=frenchblue,  postaction={}] coordinates {
            (Zero-Shot MMLU, 50.3)
        };
        \addplot[ybar, fill=babyblue,  postaction={}] coordinates {
            (Zero-Shot MMLU, 41.4)
        };  
        \addplot[ybar, fill=classicrose,  postaction={}] coordinates {
            (Zero-Shot MMLU, 46.1)
        };  
        \addplot[ybar, fill=beige,  postaction={}] coordinates {
            (Zero-Shot MMLU, 35.6)
        };  
        \addplot[ybar, fill=gray!35,  postaction={}] coordinates {
            (Zero-Shot MMLU, 49.1)
        };  

        ]
      \nextgroupplot[
            ybar,
            ymin=0, ymax=80,
            ytick={0, 10, 20, 30, 40, 50, 60, 70, 80},
            major x tick style = transparent,
            bar width=12pt,
            enlarge x limits=0.25,
            symbolic x coords={Few-Shot MMLU},  
            xtick=data,  
            axis x line*=bottom,
            axis y line*=left,
            y axis line style={opacity=0},
            ytick=\empty,
                legend style={
                        at={(2.2,0.25)},
                        draw=none,
                        anchor=south,
                        column sep=1ex,
                        font=\small,
                }
            ]  
        \addplot[ybar, fill=frenchblue,  postaction={}] coordinates {
            (Few-Shot MMLU, 52.4)
        };
        \addplot[ybar, fill=babyblue,  postaction={}] coordinates {
            (Few-Shot MMLU, 34.8)
        };  
        \addplot[ybar, fill=classicrose,  postaction={}] coordinates {
            (Few-Shot MMLU, 34.1)
        };  
        \addplot[ybar, fill=beige,  postaction={}] coordinates {
            (Few-Shot MMLU, 31.1)
        };
        \addplot[ybar, fill=gray!35,  postaction={}] coordinates {
            (Few-Shot MMLU, 47.1)
        }; 
        ]
        ]
    \end{groupplot}
    \node[below,color=forestgreen,font=\small] at (0.55,4.65) {\textbf{+3.3}};
    \node[below,color=forestgreen,font=\small] at (3.45,2.48) {\textbf{+10.2}};
    \node[below,color=forestgreen,font=\small] at (6.14,2.75) {\textbf{+8.5}};
    \node[below,color=forestgreen,font=\small] at (9.1,3.38) {\textbf{+4.2}};
    \node[below,color=forestgreen,font=\small] at (12.03,3.51) {\textbf{+17.6}};
    
    \begin{customlegend}[legend columns=-1,legend style={at={(14.2,-1.0)},draw=none,column sep=1ex},legend entries={T5-XL Flan 2022,T5-XL Flan 2021,T5-XL P3++,T5-XL SNI,OPT-IML-Max 175B}]
    \addlegendimage{black,fill=frenchblue,ybar,ybar legend}
    \addlegendimage{black,fill=babyblue,ybar,ybar legend}
    \addlegendimage{black,fill=classicrose,ybar,ybar legend}
    \addlegendimage{black,fill=beige,ybar,ybar legend}
    \addlegendimage{black,fill=gray!35,ybar,ybar legend}
    \end{customlegend}
    
    \end{tikzpicture} 
    \caption{
    \textbf{Comparing public instruction tuning collections} on Held-In, Held-Out (BIG-Bench Hard \citep{suzgun2022challenging} and MMLU \citep{hendrycks2020measuring}), and Chain-of-Thought evaluation suites, detailed in \cref{app:eval}.
    All models except OPT-IML-Max (175B) are T5-XL with 3B parameters. \textcolor{forestgreen}{Green text} indicates absolute improvement over the next best comparable T5-XL (3B) model.
    }
    \label{fig:flan-vs-competitors}
\end{figure}
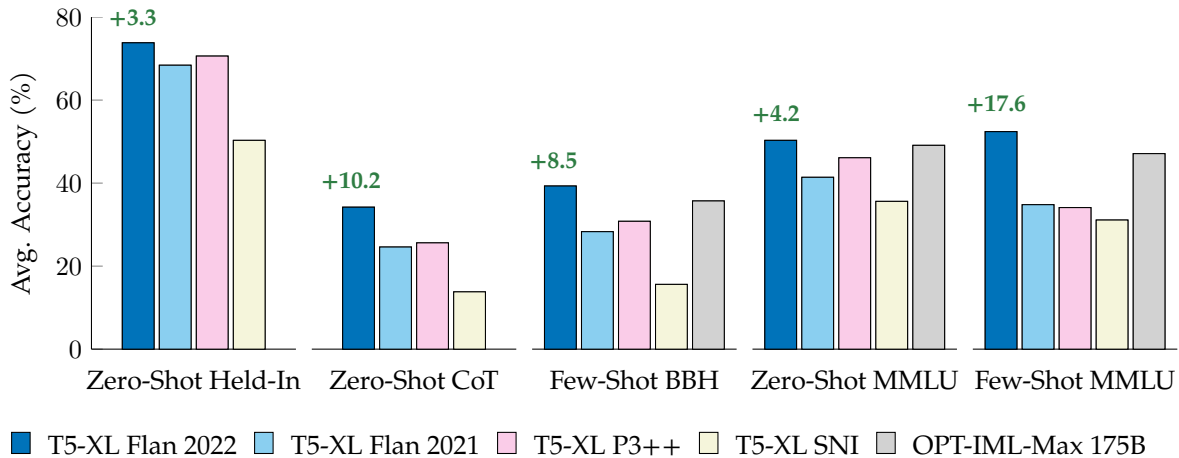

\section{Introduction}

Large language models such as PaLM \citep{chowdhery2022palm}, Chinchilla \citep{hoffmann2022training}, and ChatGPT among others \citep{brown2020language,ouyang2022training} have unlocked new capabilities in performing natural language processing (NLP) tasks from reading instructive prompts.
Prior art has shown that instruction tuning---finetuning language models on a collection of NLP tasks formatted with instructions---further enhances the ability of language models to perform an unseen task from an instruction \citep{wei2021finetuned, sanh2021multitask, min-etal-2022-metaicl}.

In this work, we evaluate the methods and results of \emph{open sourced} instruction generalization efforts, comparing their finetuning techniques and methods.
And in particular, we identify and evaluate the critical methodological improvements in the ``\flantwo{} Collection'', which is the term we use for the collection \emph{of data and methods for data augmentation and instruction tuning}, first implemented and used in \citet{chung2022scaling}.
Where \citet{chung2022scaling} focuses on the emergent and state-of-the-art results of combining Flan 2022 with PaLM 540B, this work focuses in on the details of the instruction tuning methods themselves, ablating individual factors, and comparing them directly to prior work by keeping the pretrained model size and checkpoint consistent.

The \flantwo{} Collection offers the most extensive publicly available set of tasks and methods for instruction tuning, which we have compiled in one place.
We have also supplemented this with hundreds more of our own high-quality templates, richer formatting patterns, and data augmentations.
We show that a model trained on this collection outperforms other public collections on all tested evaluation benchmarks, including the original Flan 2021 \citep{wei2021finetuned}, T0++ \citep{sanh2021multitask}, Super-Natural Instructions \citep{wang2022benchmarking}, and the concurrent work on OPT-IML \citep{iyer2022optiml}.
As shown in \cref{fig:flan-vs-competitors}, this includes 4.2\%+ and 8.5\% improvements on the MMLU \citep{hendrycks2020measuring} and BIG-Bench Hard \citep{suzgun2022challenging} evaluation benchmarks respectively, for equally sized models.

Analysis of the \flantwo{} method suggests the strong results stem both from the larger and more diverse set of tasks, but also from a set of simple finetuning and data augmentation techniques.
In particular, training on a mix of examples templatized with zero-shot, few-shot, and chain-of-thought prompts improves performance in every one of these settings, together.
For instance, adding just 10\% few-shot prompts improves zero-shot prompting results by 2\%+.
Additionally, enriching task diversity by inverting input-output pairs, as used in \citep{sanh2021multitask,min-etal-2022-metaicl}, along with balancing task sources, are both shown to be critical to performance.
The resulting Flan-T5 model converges faster and at a higher performance than T5 models in single-task finetuning---suggesting instruction-tuned models offer a more computationally-efficient starting checkpoint for downstream applications, corroborating \citet{aribandi2021ext5} and \citet{tfew2022}.

We hope making these findings and resources publicly available will unify resources around instruction tuning and accelerate research into more general-purpose language models.
We summarize this work's core contributions as follows:
\begin{itemize}\itemsep0em
    \vspace{-2 mm}
    \item Methodological: Show that training with mixed zero- and few-shot prompts yields much better performance in \textbf{both} settings (\cref{sec:mtft-zs-fs}).
    \item Methodological: Measure and demonstrate the critical techniques to effective instruction tuning: scaling \cref{sec:mtft-scaling}, enriching task variety with input inversion (\cref{sec:mtft-input-inversion}), adding chain-of-thought training data, and balancing different data sources (\cref{sec:mtft-mix-balance}).
    \item Results: Demonstrate these technical choices yield 3-17\% Held-Out task improvements over existing open source instruction tuning collections (\cref{fig:flan-vs-competitors}).
    \item Results: Demonstrate Flan-T5 serves as a stronger and more computationally-efficient starting checkpoint for single-task finetuning (\cref{sec:single-target-ft}).
    \item Open source the new \flantwo{} task collection, templates, and methods for public research.
\end{itemize}

\section{Public Instruction Tuning Collections}
\label{sec:public-collections}

\begin{figure}[h]
    \centering
    \includegraphics[width=0.99\linewidth]{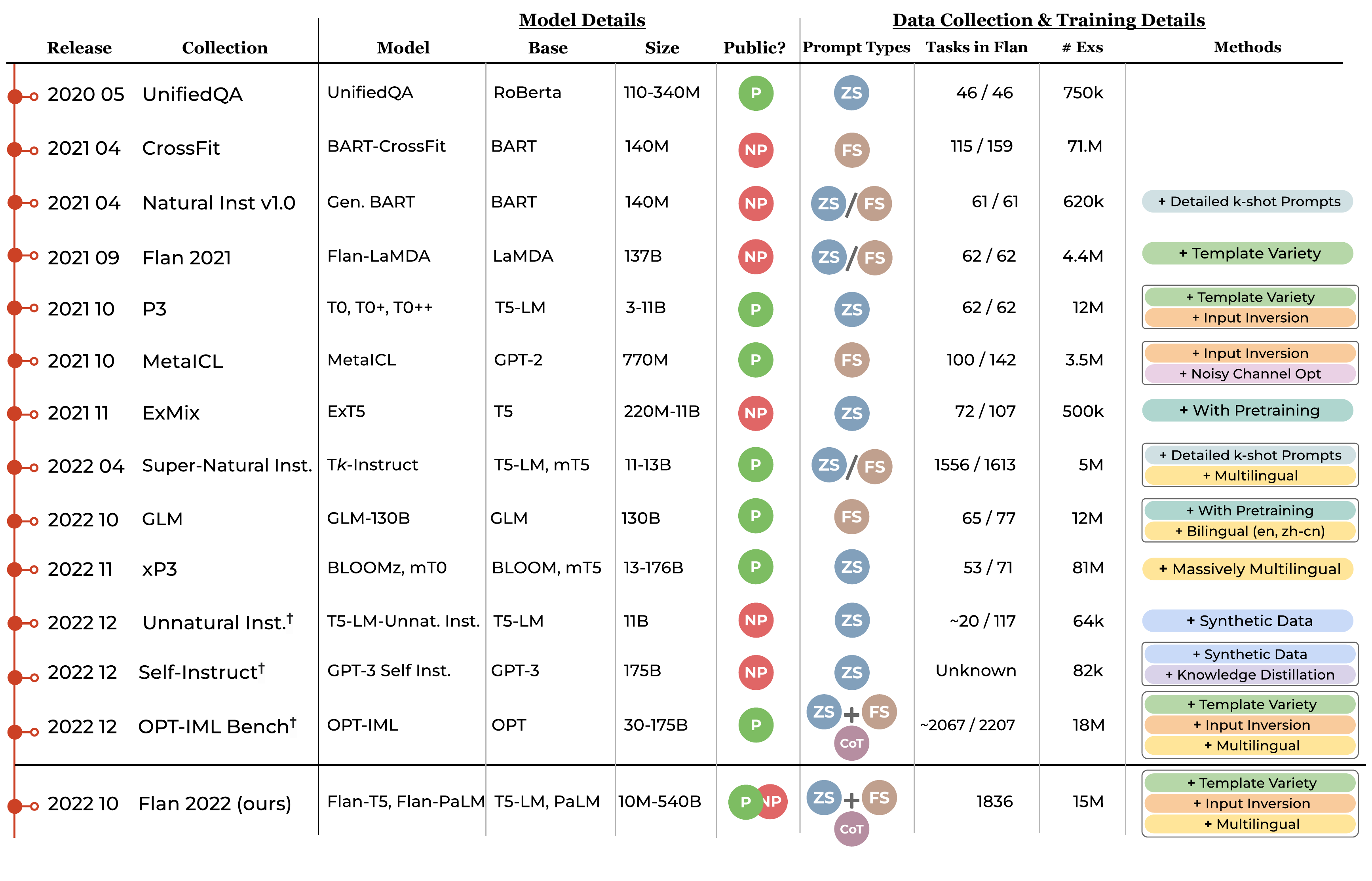}
    \caption{
    \small
    A \textbf{Timeline of Public Instruction Tuning Collections} specifies the collection release date, detailed information on the finetuned models (the base model, their size, and whether the model itself is Public (\textcolor{forestgreen}{P}) or Not Public (\textcolor{red}{NP})), what prompt specification they were trained for (zero-shot, few-shot, or Chain-of-Thought), the number of tasks contained in the Flan 2022 Collection (released with this work), and core methodological contributions in each work.\\
    Note that the number of tasks and of examples vary under different assumptions and so are estimates. For instance, the definition of ``task'' and ''task category'' vary by work, and are not easily simplified to one ontology. The reported counts for the number of tasks are reported using task definitions from the respective works.\\
    \textsuperscript{\textdagger} indicates concurrent work.\\
    }
    \vspace{-3mm}
    \label{fig:instruction-tuning-collections}
\end{figure}

\paragraph{Large Language Models} Instruction tuning has emerged as a tool to make large language models (LLMs) and their abilities more useful for interactive dialog and functional tasks.
Previous work \citep{raffel2020exploring, liu2019multi,aghajanyan-etal-2021-muppet, aribandi2021ext5} experimented with large scale multi-task finetuning, to improve downstream single target finetuning, but without instruction prompts.
UnifiedQA and others \citep{khashabi-etal-2020-unifiedqa, mccann2018natural, keskar2019unifying} unified a wide range of NLP tasks into a single generative question answering format, using prompt instructions for multi-task finetuning and evaluation.

\vspace{-2mm}
\paragraph{The First Wave}
Since 2020, several instruction tuning task collections have been released in rapid succession, outlined in \cref{fig:instruction-tuning-collections}.
Natural Instructions \citep{mishra2021cross}, Flan 2021 \citep{wei2021finetuned}, P3 (the Public Pool of Prompts, \citealp{bach-etal-2022-promptsource}) aggregated large NLP task collections and templatized them with instructions (\emph{zero-shot prompting}), specifically for finetuning models to generalize to unseen instructions. 
MetaICL \citep{min-etal-2022-metaicl} also consolidated other task collections \citep{ye2021crossfit,khashabi-etal-2020-unifiedqa} to train models to learn tasks ``in-context'' -- from several input-output examples, known as \emph{few-shot prompting}, but in this case without instructions.
Each of these works affirmed the scaling benefits of task and template diversity, and some reported strong benefits from inverting the inputs and outputs in templates to produce new tasks (``noisy channel'' in \citealp{min-etal-2022-metaicl}).

\vspace{-2mm}
\paragraph{The Second Wave}
A second wave of instruction tuning collections expanded prior resources: combining more datasets and tasks into one resource, like Super-Natural Instructions \citep{wang2022benchmarking} or OPT-IML \citep{iyer2022optiml}, adding multilingual instruction tuning in xP3 \citep{muennighoff2022crosslingual}, and Chain-of-Thought training prompts in \flantwo{} \citep{chung2022scaling}.
Both the Flan Collection and OPT-IML contain most tasks represented in prior collections.\footnote{Note that each work defines datasets, tasks, and task categories differently. For simplicity, we use their own definitions in \cref{sec:public-collections}.}
Our work is positioned here, coalescing most of these collections (of collections) and their methods, as the strongest starting point for future open source work.

\vspace{-2mm}
\paragraph{New Directions}
Concurrent and future work is beginning to explore two new directions: (a) expanding task diversity even more aggressively with synthetic data generation, particularly in creative, and open-ended dialogue \citep{selfinstruct2022, honovich2022unnatural, ye2022guess, gupta2022improving}, and (b) offering human feedback signals on model responses \citep{ouyang2022training, glaese2022improving, bai2022training, nakano2021webgpt, bai2022constitutional}.
We view most of these new directions as likely additive to a foundation of instruction tuning methods.

\vspace{-2mm}
\paragraph{Tuning with Human Feedback}
Instruction tuning on human feedback has demonstrated strong results on open-ended tasks, but at the expense of performance on a wide array of more traditional NLP tasks \citep{ouyang2022training, glaese2022improving, bai2022training, nakano2021webgpt}.
(See \citet{ouyang2022training}'s discussion of the ``alignment tax''.)
Our work focuses specifically on instruction generalization, without human feedback, for two reasons.
First, human feedback datasets are far less publicly available than instruction tuning datasets (and may be model-specific).
Second, by itself, instruction generalization shows great promise in enhancing human preferred responses on open-ended tasks, as well as improving traditional NLP metrics \citep{chung2022scaling}.
The extent of obtainable progress \emph{without} expensive human response demonstrations or ratings remains an open question, and an important pursuit to narrow the gap between public and non-public research.

\paragraph{The Importance of Open Source}

High profile research is increasingly driven by non-public data, as in the case of GPT-3 and others \citep{ouyang2022training, glaese2022improving}.
The inaccessibility of these resources inhibits the research community's ability to analyze and improve these methods in the public domain.
We narrow our purview to open source and accessible data collections, motivated by the goal of democratizing accessibility to research.

\section{Flan 2022 Instruction Tuning Experiments}
\label{sec:flan-it-exps}

Recent research has yet to coalesce around a unified set of techniques, with different tasks, model sizes, and target input formats all represented. 
We open source a new collection, first introduced in \citet{chung2022scaling}, denoted ``\flantwo{}'', which combines Flan 2021, P3++\footnote{``P3++'' is our notation for all datasets in the Public Pool of Prompts (P3): \url{https://huggingface.co/datasets/bigscience/P3}}, Super-Natural Instructions, with some additional reasoning, dialog, and program synthesis datasets.
We defer to \citet{chung2022scaling} for details of templatization and collection; and in this work we take a deeper look at key methodological improvements and compare the collection on equivalent model sizes to existing collections.

In this section, we evaluate the design decisions in Flan and discuss four in particular that yield strong improvements to the instruction tuning recipe.
These design components, outlined in \cref{sec:public-collections}, are: \textbf{(I)} using mixed zero-shot, few-shot, and Chain-of-Thought templates at training (\cref{sec:mtft-zs-fs}), \textbf{(II)} scaling T5-sized models to 1800+ tasks (\cref{sec:mtft-scaling}), \textbf{(III)} enriching tasks with input inversion (\cref{sec:mtft-input-inversion}), and \textbf{(IV)} balancing these task mixtures (\cref{sec:mtft-mix-balance}).
In \cref{sec:ablations}, we begin by measuring the value of each component and compare the final model against alternative instruction tuning collections (and their methods).

\vspace{-3mm}
\paragraph{Experimental Setup}
We finetune on the prefix language model adapted T5-LM \citep{lester-etal-2021-power}, using the XL (3B) size for all models for consistency, unless otherwise stated.
While other sizes of Flan-T5 are available, we felt XL was appropriately sized to run large-scale systematic ablations, while being sufficiently large to draw general conclusions.
We evaluate on (a) a suite of 8 ``Held-In'' tasks represented within the 1800+ training task collection (4 question answering and 4 natural language inference validation sets), (b) Chain-of-Thought (CoT) tasks (5 validation sets), and (c) the MMLU \citep{hendrycks2020measuring} and BBH \citep{suzgun2022challenging} benchmarks as our set of ``Held-Out'' tasks, as they are not included as part of Flan 2022 finetuning.
The Massivley Multitask Language Understanding benchmark (MMLU) broadly tests reasoning and knowledge capacity across 57 tasks in the sciences, social sciences, humanities, business, health, among other subjects.
BIG-Bench Hard (BBH) includes 23 challenging tasks from BIG-Bench \citep{bigbench} where \palm{} under-performs human raters.
In our ablations, we also evaluate BBH with Chain-of-Thought inputs, following \citet{chung2022scaling}.
Additional finetuning and evaluation details are provided in \cref{sec:app-setup}.

\subsection{Ablation Studies}
\label{sec:ablations}

\cref{tab:ablations} summarizes the mean contribution to Held-in, Held-out, and Chain-of-thought tasks, by individually deducting methods: mixture weight balancing (``- Mixture Balancing"), Chain-of-thought tasks (``- CoT"), mixed prompt settings (``- Few Shot Templates"), and Input Inversion (``- Input Inversion").
Flan-T5 XL leverages all four of these methods together.
We also finetune T5-XL-LM on other collections, including Flan 2021, P3++, Super-Natural Instructions for comparison.

\begingroup
\setlength{\tabcolsep}{4pt}
\begin{table*}[ht]
    \centering
    \small
    \begin{tabular}{l | ccccc}
    \toprule
    \textsc{Model} & \textsc{Held-In} & \textsc{CoT} & \textsc{MMLU} & \textsc{BBH} & \textsc{BBH-CoT} \\
    \midrule
    T5-XL \flantwo{} & \textbf{73.8 / 74.8} & 35.8 / \textbf{34.1} & \textbf{50.3 / 52.4} & 26.2 / \textbf{39.3}  & \textbf{33.9 / 35.2} \\
    \midrule
    - CoT & 73.3 / 73.2 & 28.8 / 24.6 & 47.5 / 46.9 & 18.2 / 30.0 & 18.2 / 12.0 \\
    - Input Inversion & \textbf{73.8} / 74.1 & 32.2 / 23.5 & 41.7 / 41.2 & 18.4 / 24.2	& 15.7 / 13.0 \\
    - Mixture Balancing & 71.2 / 73.1 & 32.3 / 30.5 & 45.4 / 45.8 & 15.1 / 24.3 & 13.8 / 15.4 \\
    - Few Shot Templates & 72.5 / 62.2 & \textbf{38.9}	/ 28.6 & 47.3 / 38.7 & 27.6 / 30.8 & 18.6 / 23.3 \\
    \midrule
    T5-XL Flan 2021 & 68.4 / 56.3 & 24.6 / 22.7 &	41.4 / 34.8 & \textbf{28.1} / 28.3 & 26.0 / 26.9 \\
    T5-XL P3++ & 70.5 / 62.8 & 25.6 / 25.6 & 46.1 / 34.1 & 26.0 / 30.8 & 23.4 / 26.1 \\
    T5-XL Super-Natural Inst. & 50.3 / 42.2&	13.8 / 14.3 & 35.6 / 31.1 & 10.4 / 15.6 & 8.0 / 12.5 \\
    GLM-130B\textsuperscript{\textdagger} & - &	- & -- / 44.8 & - & - \\
    OPT-IML-Max 30B\textsuperscript{\textdagger} & - &	- & 46.3 / 43.2 &  -- / 30.9 & - \\
    OPT-IML-Max 175B\textsuperscript{\textdagger} & - &	- & 49.1 / 47.1 &  -- / 35.7 & - \\
    \midrule
    \flantwo{} - Next Best T5-XL & \textcolor{forestgreen}{+3.3} / \textcolor{forestgreen}{+12} & \textcolor{forestgreen}{+10.2} / \textcolor{forestgreen}{+8.5} & \textcolor{forestgreen}{+4.2} / \textcolor{forestgreen}{+17.6} & \textcolor{color3}{-1.9} / \textcolor{forestgreen}{+8.5} & \textcolor{forestgreen}{+7.9} / \textcolor{forestgreen}{+8.3}\\
    \bottomrule
    \end{tabular}
    \caption{
    \textbf{Method Ablations (top)} show the importance of each method for Flan-T5 XL.
    \textbf{Collection Ablations (bottom)} evaluate Flan-T5 XL against T5-XL finetuned on other instruction tuning collections: FLAN 2021, P3++, and Super-Natural Instructions.
    \textbf{\flantwo{} - Next Best T5-XL} shows the improvement of Flan-T5 XL over the next best T5-XL (comparatively sized) finetuned on another collection.
    Metrics are reported in both zero-shot / few-shot settings across Held-In, Chain-of-Thought, and Held-Out (MMLU, BBH) tasks. \\
    \textsuperscript{\textdagger} We also inlcude the results reported by OPT-IML \citep{iyer2022optiml} and GLM-130B \citep{zeng2022glm}.
    }
    \label{tab:ablations}
\end{table*}
\endgroup

Each of the ablated components of Flan contributes improvements to different metrics: Chain-of-Thought training to Chain-of-Thought evaluation, input inversion to Held-Out evaluations (MMLU and BBH), few-shot prompt training to few-shot evaluations, and mixture balancing to all metrics.

As compared to T5-XL models trained on alternative instruction tuning collections (and their methods), Flan outperforms in almost every setting.
While previous collections are tuned specifically to zero-shot prompts, Flan-T5 XL is tuned for either zero- or few-shot prompts.
This yields performance margins of +3-10\% for most of the zero-shot settings, and margins of 8-17\% for the few-shot settings.
Most impressively, \flantwo{} outperforms OPT-IML-Max's much larger (10x) 30B and (58x) 175B models.
Next, we isolate some of \flantwo{}'s ablated methods individually, to examine the benefits of each.

\subsection{Training with Mixed Prompt Settings}
\label{sec:mtft-zs-fs}
Prior work has shown a wide variety of input templates per task can improve performance.
However, separate from the wording of the instruction template, these prior LLMs mostly tune with template sets \emph{targeted to a single prompt setting}: for zero-shot prompting \citep{wei2021finetuned, sanh2021multitask, aghajanyan-etal-2021-muppet,aribandi2021ext5} or for few-shot prompting \citep{min-etal-2022-metaicl, wang2022benchmarking}.

An underappreciated design decision in InstructGPT \citep{ouyang2022training} was to mix training templates for each of these prompt settings, rather than target a single setting.
However, since \citet{ouyang2022training} do not examine this choice, we expected a performance trade-off in finetuning for zero-shot or few-shot prompting performance -- particularly for smaller models.
Instead, we find training with mixed zero- and few-shot prompts significantly improves performance in \textbf{both} settings -- most surprisingly, even for models with only 3B parameters.

\makeatletter
\newenvironment{customlegend2}[1][]{%
    \begingroup
    \pgfplots@init@cleared@structures
    \pgfplotsset{#1}%
}{%
    \pgfplots@createlegend
    \endgroup
}%

\def\addlegendimage{\csname pgfplots@addlegendimage\endcsname}

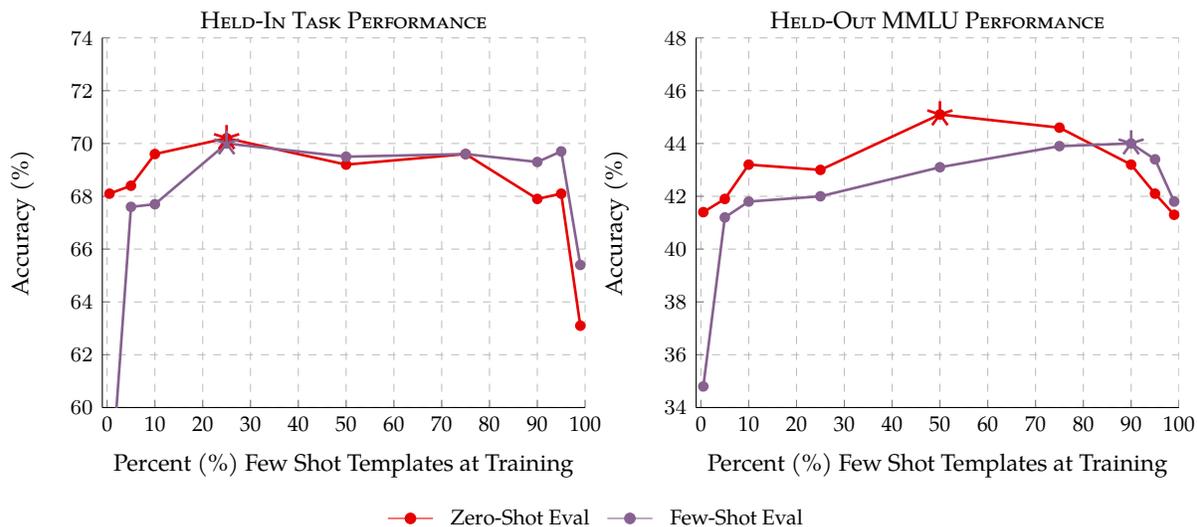
\begin{figure}[ht]
    \begin{centering}
    \begin{tikzpicture}
        \pgfplotsset{footnotesize,samples=10}
        \begin{groupplot}[
            group style = {group size = 2 by 1, horizontal sep = 42pt},
            width = 8cm, 
            height = 6.5cm]
            \nextgroupplot[
                align = center,
                title = {\textsc{Held-In Task Performance}},
                legend style={at={(-0.12,1.4)},anchor=south},
                xmin=-0.01, xmax=1.0,
                ymin=60, ymax=74,
                xtick={0.0, 0.1, 0.2, 0.3, 0.4, 0.5, 0.6, 0.7, 0.8, 0.9, 1.0},
                axis x line*=bottom,
                axis y line*=left,
                xticklabels={0, 10, 20, 30, 40, 50, 60, 70, 80, 90, 100},
                xlabel={Percent (\%) Few Shot Templates at Training},
                ylabel={Accuracy (\%)},
                ytick={60, 62, 64, 66, 68, 70, 72, 74},
                grid style=dashed,
                x label style={at={(axis description cs:0.5,-0.1)},anchor=north},
                y label style={at={(axis description cs:-0.12,0.5)},anchor=south},
                xtick pos=bottom,
                ytick pos=left,
                grid=both,
                ]
                \addplot[ 
                    color=color4,
                    mark=*,
                    mark size=1.5pt,
                    line width=1pt,
                    ]
                    coordinates {
                    (0.005, 68.1)
                    (0.05, 68.4)
                    (0.1, 69.6)
                    (0.25, 70.2)
                    (0.5, 69.2)
                    (0.75, 69.6)
                    (0.9, 67.9)
                    (0.95, 68.1)
                    (0.99, 63.1)
                    };
                \addplot[ 
                    color=frenchlilac,
                    mark=*,
                    mark size=1.5pt,
                    line width=1pt,
                    ]
                    coordinates {
                    (0.005, 56.3)
                    (0.05, 67.6)
                    (0.1, 67.7)
                    (0.25, 70)
                    (0.5, 69.5)
                    (0.75, 69.6)
                    (0.9, 69.3)
                    (0.95, 69.7)
                    (0.99, 65.4)
                    };

                \addplot[ 
                    color=color4,
                    mark=star,
                    mark size=5pt,
                    line width=1pt,
                    ]
                    coordinates {
                    (0.25, 70.2)
                    };
                \addplot[ 
                    color=frenchlilac,
                    mark=star,
                    mark size=5pt,
                    line width=1pt,
                    ]
                    coordinates {
                    (0.25, 70)
                    };

            \nextgroupplot[
                align = center,
                title = {\textsc{Held-Out MMLU Performance}},
                legend style={at={(-0.12,1.4)},anchor=south},
                xmin=-0.01, xmax=1.0,
                ymin=34, ymax=48,
                xtick={0.0, 0.1, 0.2, 0.3, 0.4, 0.5, 0.6, 0.7, 0.8, 0.9, 1.0},
                axis x line*=bottom,
                axis y line*=left,
                xticklabels={0, 10, 20, 30, 40, 50, 60, 70, 80, 90, 100},
                xlabel={Percent (\%) Few Shot Templates at Training},
                ylabel={Accuracy (\%)},
                ytick={34, 36, 38, 40, 42, 44, 46, 48, 50},
                grid style=dashed,
                x label style={at={(axis description cs:0.5,-0.1)},anchor=north},
                y label style={at={(axis description cs:-0.12,0.5)},anchor=south},
                xtick pos=bottom,
                ytick pos=left,
                grid=both,
                ]
                
                \addplot[ 
                    color=color4,
                    mark=*,
                    mark size=1.5pt,
                    line width=1pt,
                    ]
                    coordinates {
                    (0.005, 41.4)
                    (0.05, 41.9)
                    (0.1, 43.2)
                    (0.25, 43.0)
                    (0.5, 45.1)
                    (0.75, 44.6)
                    (0.9, 43.2)
                    (0.95, 42.1)
                    (0.99, 41.3)
                    };
                \addplot[ 
                    color=frenchlilac,
                    mark=*,
                    mark size=1.5pt,
                    line width=1pt,
                    ]
                    coordinates {
                    (0.005, 34.8)
                    (0.05, 41.2)
                    (0.1, 41.8)
                    (0.25, 42)
                    (0.5, 43.1)
                    (0.75, 43.9)
                    (0.9, 44)
                    (0.95, 43.4)
                    (0.99, 41.8)
                    };

                \addplot[ 
                    color=color4,
                    mark=star,
                    mark size=5pt,
                    line width=1pt,
                    ]
                    coordinates {
                    (0.5, 45.1)
                    };
                \addplot[ 
                    color=frenchlilac,
                    mark=star,
                    mark size=5pt,
                    line width=1pt,
                    ]
                    coordinates {
                    (0.9, 44)
                    };
        \end{groupplot}
        
    \begin{customlegend2}[legend columns=-1,legend style={at={(9.5,-1.2)},draw=none,column sep=1ex},legend entries={Zero-Shot Eval, Few-Shot Eval}]
    \addlegendimage{color4,fill=color4,mark=*,sharp plot}
    \addlegendimage{frenchlilac,fill=frenchlilac,mark=*,sharp plot}
    \end{customlegend2}
    
    \end{tikzpicture}
    \caption{
    \textbf{Training jointly with zero-shot and few-shot prompt templates improves performance} on both Held-In and Held-Out tasks.
    The stars indicate the peak performance in each setting.
    } 
    \label{fig:zero-few-shot}
    \end{centering}
\end{figure}

\cref{fig:zero-few-shot} shows (1) adding as little as 5\% few-shot training templates can dramatically improve zero-shot performance, and (2) adding 10\%+ of zero-shot data improves few-shot performance too.
Both Held-In and Held-Out tasks peak anywhere between 10-90\% of few-shot data, but this range is consistently higher than training with only one prompt setting.

\subsection{Scaling Small Models to 1.8k+ Tasks}
\label{sec:mtft-scaling}

The most recent and concurrent publicly available instruction tuning efforts, like \flantwo{}, train on thousands of tasks \citep{wang2022benchmarking,iyer2022optiml}, but operate on different task compositions and underlying training methods.
To measure the impact of scaling model sizes and tasks for the  \flantwo{} collection, we finetune T5-LM adapted models (Small, Base, Large, XL, XXL) on randomly selected task subsets (8, 25, 50, 100, 200, 400, 800, all 1873). 
Every finetuning run is guaranteed to include the Held-In tasks, so we can estimate how task scaling impacts the model capacity to maintain performance on a given task its already seen.

\input{fables/scaling-laws}

\cref{fig:scaling-laws} demonstrates that both Held-In and Held-Out tasks appear to benefit from adding hundreds of finetuning tasks.
Held-in task evaluations peak around 200 total tasks, and diminish in performance as more tasks are added, though larger models peak later and diminish less.
Held-out task performance increases log-linearly with the number of tasks, achieving the highest performances with all 1836 tasks.
Surprisingly, only T5-Small appears to exceed its Held-Out task performance before 1836 tasks, while larger model sizes continue to improve.
These results suggest (a) even T5-Base may not have exhausted its capacity with thousands of tasks, and (b) the largest LMs could benefit from thousands more tasks for Held-In and Held-Out task performance.

One necessary assumption of this analysis is that all tasks are defined and counted equally.
\cref{sec:mtft-mix-balance} demonstrates how not all task sources are equally beneficial to training, and the model performance may saturate from too many tasks from one source (e.g. Super-Natural Instructions).
We would caution conclusions that task scaling beyond 1800 would translate to increased returns without also paying attention to task diversity and quality.

\subsection{Task Enrichment with Input Inversion}
\label{sec:mtft-input-inversion}

Prior instruction tuning work has enriched their diversity of tasks by inverting the ($x$, $y$) input-output pairs in supervised tasks---referred to as ``prompts not intended for the original task'' in P3 \citep{bach-etal-2022-promptsource} or the ``noisy channel'' in MetaICL \citep{min-etal-2022-metaicl}.
For example, a dataset may be originally designed for, given a question $x$, evaluate if a model can answer $y$. Input inversion instead gives a model the answer $y$ and trains it to generate the question $x$.    
This is an easy method to enrich the task variety given a limited set of data sources.
However, it isn't clear that this method remains helpful when 100s of unique data sources and 1000s of tasks are already available.

To assess this, we enrich our mixtures with input inverted tasks (details and examples in \cref{sec:app-input-inversion}) and measure the effect.
In \cref{tab:ablations} we find this is not beneficial for Held-In performance, but strongly beneficial for Held-Out performance.
These benefits invigorate the prospect of data augmentation techniques for LLM finetuning, which had previously been shown to have diminishing returns the longer models are pretrained \citep{longpre2020effective}.

\begin{table}[ht]
    \centering
    \begin{tabular}{l | ccc}
    \toprule
    \textsc{Train Mixtures} &  & \textsc{Metrics} & \\
    & Held-In & CoT & MMLU \\
    \midrule
    All (Equal) & 64.9 & 41.4 & 47.3 \\
    \midrule
    All - Flan 2021 & 55.3 & 38.6 & 45.7 \\
    All - T0-SF & 63.2 & \textbf{43.4} & 44.7 \\
    All - Super-Nat. Inst. & 65.9 & 42.2 & 46.8 \\
    All - CoT & 65.6 & 29.1 & 46.8 \\
    All - Prog. Synth. & 66.9 & 42.3 & 46.8  \\
    All - Dialog & 65.4 & 40.3 & 47.1 \\
    \midrule
    All (Weighted) & \textbf{66.4} & 40.1 & \textbf{48.1} \\
    \bottomrule
    \end{tabular}
    \caption{
    Subsets of tasks are left out from an equally weighted mixture to measure their importance.
    \textbf{T0-SF and Flan 2021 finetuning are most important for MMLU, while Chain-of-Thought (CoT) finetuning is most important for Chain-of-Thought evaluation.}}
    \label{tab:mixture-ranking}
\end{table}

\subsection{Balancing Data Sources}
\label{sec:mtft-mix-balance}

Scaling architecture size and the number of tasks are effective, but our results suggest the mixture weighting deserves as much attention to optimize results. 
To converge on a balanced weighting, we omit different sets of task sources, one at a time (Flan 2021, \tzeromixture{}, Super-Natural Instructions, Chain-of-Thought, Dialog, and Program Synthesis), and rank their contributions on the MMLU benchmark.\footnote{Following \citet{chung2022scaling} we refer to the subset of P3++ that is not in Flan 2021 as T0-SF (SF stands for “sans Flan”).}.

As shown in \cref{tab:mixture-ranking}, Flan 2021 and \tzeromixture{} are among the most beneficial mixtures, followed by Super-Natural Instructions and Chain-of-Thought, with Dialog and Program Synthesis last.
These findings are corroborated by \citet{iyer2022optiml} who extensively test data mixing proportions, and also determine their Flan 2021, \tzeromixture{}, and T5 mixtures are the most broadly beneficial.
Additionally, they find Super-Natural Instructions has limited scaling benefits on Held-Out task performance, which they relate to its unique input format and instruction design.
Notably, Chain-of-thought finetuning appears beneficial across all our evaluation settings, especially considering they contain far fewer tasks than Flan 2021, \tzeromixture{} or Natural Instructions.

\input{fables/single-target-finetuning} 

We used these findings to significantly narrow the mixture weights search space, and used our practitioner's intuition from there.
This strategy is simple but effective, as shown in \cref{tab:ablations}, but leaves ample room for more sophisticated future work.

\input{fables/single-target-convergence}

\subsection{Discussion}
\label{sec:discussion}

OPT-IML \citep{iyer2022optiml} presents the closest comparison to this work, including a similar collection of tasks, examples and techniques.
However, while their used tasks are all publicly sourced, their collection, with templates, processing, and example mixing, is not released, and as a result cannot be easily compared.
\citet{iyer2022optiml} report that Flan-T5-XL (3B) and XXL (11B) outperforms OPT-IML-Max 175B on both MMLU and BBH.
As they discuss, these differences may arise from any combination of pre-training, model architecture, and instruction tuning.
Model architecture and pretraining before instruction tuning can play a significant role \citep{wang2022language}.
But there are many other details in instruction tuning that may vary between \flantwo{} and OPT-IML.
Likely candidates are are: example templatization, how the mixed input prompting procedures are used at training, and task composition.

How significant are each of these difference?
While OPT-IML contains more tasks than \flantwo{}, we estimate approximately $94\% (2067 / 2207)$ are also used in the \flantwo{} collection\footnote{This is calculated using their definition of ``task'' (reported in \citet{iyer2022optiml}'s Table 1), which does not deduplicate across collections.}, and very few tasks in \flantwo{} are not contained in some format in OPT-IML.
This suggests the overall difference in task diversity is not significant when using a shared definition of ``task''.
Task mixture rates also emphasize similar sources, including Flan 2021 (46\% vs 20\%), PromptSource/P3 (28\% vs 45\%), and Super-Natural Instructions (25\% vs 25\%), for \flantwo{} and OPT-IML respectively.\footnote{Note that 46\% weight for \flantwo{} is actually on Muffin from \citet{chung2022scaling} which combines Flan 2021 with new dialog and program synthesis tasks.}
OPT-IML's other collections (Crossfit, ExMix, T5, U-SKG) are not weighted significantly: 4\%, 2\%, 2\%, 2\% respectively.

We believe example templatization and the mixed prompt formats may pose the largest differences with OPT-IMLs instruction tuning.
Our template repository was significantly updated from Flan 2021, adding variety not just in instructions, but also along dimensions.
For instance, the templatization procedure varies where the instruction is placed (before or after few-shot prompts), the spacing and separators between few-shot and Chain-of-Thought prompts, and the formatting permutations of answer options (and their targets) for multiple-choice examples, which sometimes includes and sometimes excludes answer options in the inputs or exemplars.
While we do not have dedicated experiments comparing many iterations of development, we found these procedures dramatically augment input variety and showed repeated performance improvements.
Our example templatizing procedure is open sourced for inspection and future work.

\section{Instruction Tuning Enhances Single-Task Finetuning}
\label{sec:single-target-ft}

In applied settings, machine learning practitioners deploy NLP models finetuned (FT) specifically for a single target task, usually where finetuning data is already available. While prior work has shown the benefits of intermediate finetuning~\citep{pruksachatkun2020intermediate,vu-etal-2020-exploring} or multi-task finetuning~\citep{aghajanyan-etal-2021-muppet,aribandi2021ext5} 
for downstream tasks, this has not been studied extensively for instruction-tuned models.

We evaluate \flantwo{} instruction tuning as an intermediary step before single target finetuning, to understand if Flan-T5 would serve as a better starting checkpoint for applied practitioners.
We evaluate three settings in \cref{fig:single-target}: finetuning T5 directly on the target task as the conventional baseline (blue bars), using Flan-T5 without further finetuning (beige bars), and finetuning Flan-T5 further on the target task (red bars).

\vspace{-3mm}
\paragraph{Pareto Improvements to Single Task Finetuning}
For both sets of Held-In and Held-Out tasks examined, finetuning Flan-T5 offers a pareto improvement over finetuning T5 directly. In some instances, usually where finetuning data is limited for a task, Flan-T5 without further finetuning outperforms T5 with task finetuning.

\vspace{-3mm}
\paragraph{Faster Convergence \& Computational Benefits}
Using Flan-T5 as a starting checkpoint has an added benefit in training efficiency.
As demonstrated in \cref{fig:single-target-convergence}, Flan-T5 converges much more quickly than T5 during single target finetuning, as well as peaking at higher accuracies.
These convergence results also suggest there are strong green-AI incentives for the NLP community to adopt instruction-tuned models, like Flan-T5 for single-task finetuning, rather than conventional non-instruction-tuned models.
While instruction tuning is more computationally-expensive than single-task finetuning, it is a one-time cost.
On the contrary, pretrained models that require extensive finetuning become more costly when aggregating over many millions of additional training steps \citep{wu2022sustainable, bommasani2021opportunities}. Instruction-tuned models offer a promising solution to significantly reduce the amount of finetuning steps across a wide swathe of tasks, if they are adopted as a new standard starting point for single-task finetuning.

\section{Related Work}
\label{sec:rw}

\paragraph{Large Language Models} 
As the foundation of instruction tuning, the practice of pretraining one general-purpose language representation that is useful for multiple downstream tasks has a long tradition that goes back at least \citet{word2vec} and \citet{pretrainedLSTM}. 
In 2018, \citet{peters-etal-2018-deep} and \citet{devlin-etal-2019-bert} cemented the paradigm of pretraining a large model on a large unsupervised corpus, and the field of NLP quickly converged to using these models which substantially outperform the prior art of non-pretrained task-specific LSTM models on all tasks. 
However, the dominate way to access that high-quality syntactic and semantic knowledge encoded in pretrained models was not to prompt them with instructions,
but to train an additional task-specific linear layer that maps the model activations into numerical class labels.
A short year later, \citet{radford2019language}, \citet{raffel2020exploring}, and \citet{lewis-etal-2020-bart} popularized the notion that downstream tasks—and multiple tasks—can be jointly learned by directly using the pretrained LM head to generate the answers in natural language (cf. task-specific numerical class labels), 
the task-general nature of these generative models became the precursor to many multitask transfer learning studies \citep{mccann2018natural,khashabi-etal-2020-unifiedqa,ye2021crossfit,vu-etal-2020-exploring}, which in turn led to the first wave of instruction tuning as described in \cref{sec:public-collections}.

The continuing advancement in research on the pretraining corpora, architectures and pretraining objectives of LMs also has a large impact on instruction tuning. 
As of 2022, decoder-only left-to-right causal Transformers dominate the market of models larger than 100B \citep{brown2020language,thoppilan2022lamda,rae2021scaling,chowdhery2022palm,hoffmann2022training}, 
and all models of such size class with fully public model parameters are decoder-only \citep{gpt-j,scao2022bloom,zhang2022opt},
the decision of which are often due to better hardware and software framework support. 
However, \citet{raffel2020exploring}, \citet{lewis-etal-2020-bart}, and \citet{tay2022unifying} have consistently found that left-to-right causal language modeling is a suboptimal objective, while \citet{tay2022transcending} and \citet{wang2022language} particularly showed that a mixture of non-sequential objectives is much superior for downstream tasks with zero-shot and few-shot prompting.
An additional factor which remains under-explored is the relationship between pretraining corpora, instruction tuning, and downstream abilities.
Typically, public models are all trained on one of a few public corpora: C4 \citep{raffel2020exploring}, The Pile \citep{gao2020pile}, or ROOTs \citep{laurencconbigscience}.

\vspace{-2mm}
\paragraph{Instruction Tuning} In \cref{sec:public-collections} we outline major developments in instruction tuning. 
Other important developments include the prospect of complimenting or replacing few-shot in-context learning-the currently predominate method of evaluating pretrained and instruction-tuned models—with parameter-efficient tuning.
As standard finetuning of models larger than 100B requires a high number of accelerators with the right interconnects often too expensive even for many industry labs, parameter-efficient tuning (a.k.a. continuous or soft “prompt tuning”) shows that only updating a small subset of model parameters can reach comparable performance as fully tuning all model parameters (\citealp{lester-etal-2021-power,vu-etal-2022-spot,lora}; see \citealp{he2022towards} for a detailed analysis).
Notably, \citet{tfew2022} show that, due to the long sequence length of few-shot ICL and that the few-shot exemplars need to be repeatedly inferenced for evaluating every example, parameter-efficient tuning can be computationally cheaper and higher performing than in-context learning.
Further, \citet{tfew2022}, \citet{vu-etal-2022-spot}, \citet{wei2021finetuned}, and \citet{med-palm} collectively show that both single-task and multi-task parameter-efficient tuning can be productively combined with instruction tuning, either before or after regular full-model instruction tuning.
This line of work makes it easy for other researchers to build on top of a general-domain instruction-tuned model, 
and collect a custom instruction-tuning mixture for their use,
e.g., with multiple modalities \citep{2022_palm_saycan,huang2022inner,multimodal-inst-tuning} or special domains such as science and medicine \citep{minerva,med-palm}. 

\vspace{-2mm}
\paragraph{Problems Addressed by Instruction Tuning \& Alignment Techniques}
Instruction tuning is part of a line of work designed to ``align'' language models with more useful objectives and human preferences.
In the absence of such methods, language models are known to demonstrate toxic/harmful behaviour \citep{sheng-etal-2019-woman,liang2021towards,wallace-etal-2019-universal}, generate non-factual information \citep{maynez-etal-2020-faithfulness,longpre2021entity,devaraj-etal-2022-evaluating}, and other challenges in deployment and evaluation \citep{zellers2019defending,mcguffie2020radicalization,talat2022you}.
Analyzing, evaluating and mitigating these problems pose a promising direction for future work \citep{gao2022attributed,ganguli2022red}.
Instruction tuning warrants greater investigation, as it has already demonstrated itself an encouraging remedy in reducing NLP bias metrics, as shown in \citet{chung2022scaling}.

\section{Conclusions}
The new \flantwo{} instruction tuning collection unifies the most popular prior public collections and their methods, while adding new templates and simple improvements like training with mixed prompt settings.
The resulting collection outperforms Flan 2021, P3++, Super-Natural Instructions, and OPT-IML-Max 175B on Held-In QA, NLI, and Chain-of-Thought tasks, and Held-Out MMLU and BBH, often by large margins.
Results suggest this new collection serves as a more competitive starting point for researchers and practitioners interested in both generalizing to new instructions, or finetuning on a single new task.

\section*{Acknowledgements}

We would like to thank Ed H Chi, Xinyun Chen, and Colin Raffel for their advice and feedback on the paper.

\clearpage
\bibliographystyle{plainnat}
\bibliography{main}

\clearpage
\appendix
\addcontentsline{toc}{section}{Appendix} 
\part{Appendix} 
\parttoc

\section{Experimental Details}
\label{sec:app-setup}

\subsection{Instruction Tuning}

The Flan Collection experiments are assembled and run using T5X \citep{roberts2022t5x}.
Our instruction tuning follows the same setup described in \citet{chung2022scaling}.
For few-shot and few-shot Chain-of-Thought prompts during finetuning our templatizing procedure generates few-shot examples with 2, 3, or 5 exemplars.
The experiments in this work use a slightly earlier version of the \flantwo{} collection the one we are releasing, which had some minor improvements to the templates.

The mixture weights used to balance the various sources of data were informed by experiments in \cref{sec:mtft-mix-balance}, along with the resulting practitioner intuition.

\subsection{Single-Task Finetuning}

\begingroup
\setlength{\tabcolsep}{4pt}
\begin{table*}[ht]
    \centering
    \small
    \begin{tabular}{ll | cccc | r}
    \toprule
    & & \multicolumn{4}{c}{\textsc{Used in}} & \\
    \textsc{Dataset} & \textsc{Metric} &  \textsc{Held-In} & \textsc{CoT} & \textsc{ST-FT Held-In} & \textsc{ST-FT Held-Out} & \textsc{Citation} \\
    \midrule
    ARC E+C & Acc & $\checkmark$ & & $\checkmark$ & & \citep{clark2018think} \\
    ANLI R1+R2+R3 & 3-class F1 & $\checkmark$ & & $\checkmark$ & & \citep{nie2020adversarial} \\
    AI2 Mid. Science & 4-class F1 & $\checkmark$ & & $\checkmark$ & & \href{http://data.allenai.org/ai2-science-questions}{AI2 Science Questions} \\
    BoolQ & AUC-ROC & $\checkmark$ & & $\checkmark$ & & \citep{clark2019boolq} \\
    RTE & AUC-ROC & $\checkmark$ & & $\checkmark$ & & \citep{bentivogli2009fifth} \\
    SQuAD V2 & F1 & & & $\checkmark$ & & \citep{rajpurkar2018know} \\
    CosmosQA & Acc & & & $\checkmark$ & & \citep{huang2019cosmos} \\
    \midrule
    GSM8K & Acc & & $\checkmark$ & & & \citep{cobbe2021training} \\
    StrategyQA & Acc & & $\checkmark$ & & & \citep{geva2021did} \\
    SVAMP & Acc & & $\checkmark$ & & & \citep{patel2021nlp} \\
    Asdiv & Acc & & $\checkmark$ & & & \citep{miao2020diverse} \\
    CommonsenseQA & Acc & & $\checkmark$ & & & \citep{talmor2019commonsenseqa} \\
    \midrule
    WANLI & Acc & & & & $\checkmark$ & \citep{liu2022wanli} \\
    MedNLI & Acc & & & & $\checkmark$ & \citep{romanov-shivade-2018-lessons} \\
    CondaQA & Acc & & & & $\checkmark$ & \citep{ravichander2022condaqa} \\
    PubmedQA & F1 & & & & $\checkmark$ & \citep{jin-etal-2019-pubmedqa} \\
    CxC & Spearman & & & & $\checkmark$ & \citep{parekh-etal-2021-crisscrossed} \\
    \bottomrule
    \end{tabular}
    \caption{
    \textbf{Datasets used for Various Finetuning and Evaluation Experiments.} 
    ST-FT stands for Single Task Finetuning.
    }
    \label{tab:app-datasets}
\end{table*}
\endgroup

For single-task finetuning, described in \cref{sec:single-target-ft}, our models are finetuned for 100${,}$000 steps for all tasks. We use a constant learning rate of 0.001, a dropout probability of 0.1, and a batch size of 128 length-512 sequences. We save a checkpoint every 20 steps and report test performance on the model checkpoint corresponding to the highest validation performance. For tasks without a validation split, we hold out 1024 training examples for validation. For tasks without a test split, we hold out 1024 training examples for validation and report results on the original validation set. For PubmedQA, we do not use any of the unlabeled and artificially generated QA instances associated with the dataset. For CxC, we only consider the text-text portion of the dataset, following~\citet{vu-etal-2022-spot}. For tasks with less than 1K training examples, we report average results across 3 random seeds.

We also evaluate on certain metrics to account for label skew in some of the datasets, as shown in \cref{tab:app-datasets}.

\subsection{Evaluation}
\label{app:eval}

For Held-In evaluations we use the validation sets from 4 question answering (QA) tasks, BoolQ, ARC Easy, ARC Challenge, and AI2's Middle School Science Exams, and 4 natural language inference (NLI) tasks, including ANLI R1, R2, R3, and RTE. 
These datasets are contained in the \flantwo{} finetuning collection and represent challenging benchmarks, often used to evaluate LLMs on QA and NLI.
The Held-In score is the mean accuracy across these 8 tasks.

For the Chain-of-Thought (CoT) evaluation, we use the mean accuracy across 5 datasets which have been prepared with prompts which request step-by-step explanations in their target answers: GSM8K, StrategyQA, SVAMP, Asdiv, and CommonsenseQA.

For the Held-Out tasks, we use MMLU's suite of 57 exams, and BBH's suite of 23 tasks where PaLM performed worse than the average human annotators.
MMLU tasks were removed from the Super-Natural Instructions part of the \flantwo{} collection at training, to ensure they were Held-Out.

\section{Input Inversion Details}
\label{sec:app-input-inversion}

\begin{figure}[h]
    \centering
    \includegraphics[width=0.7\linewidth]{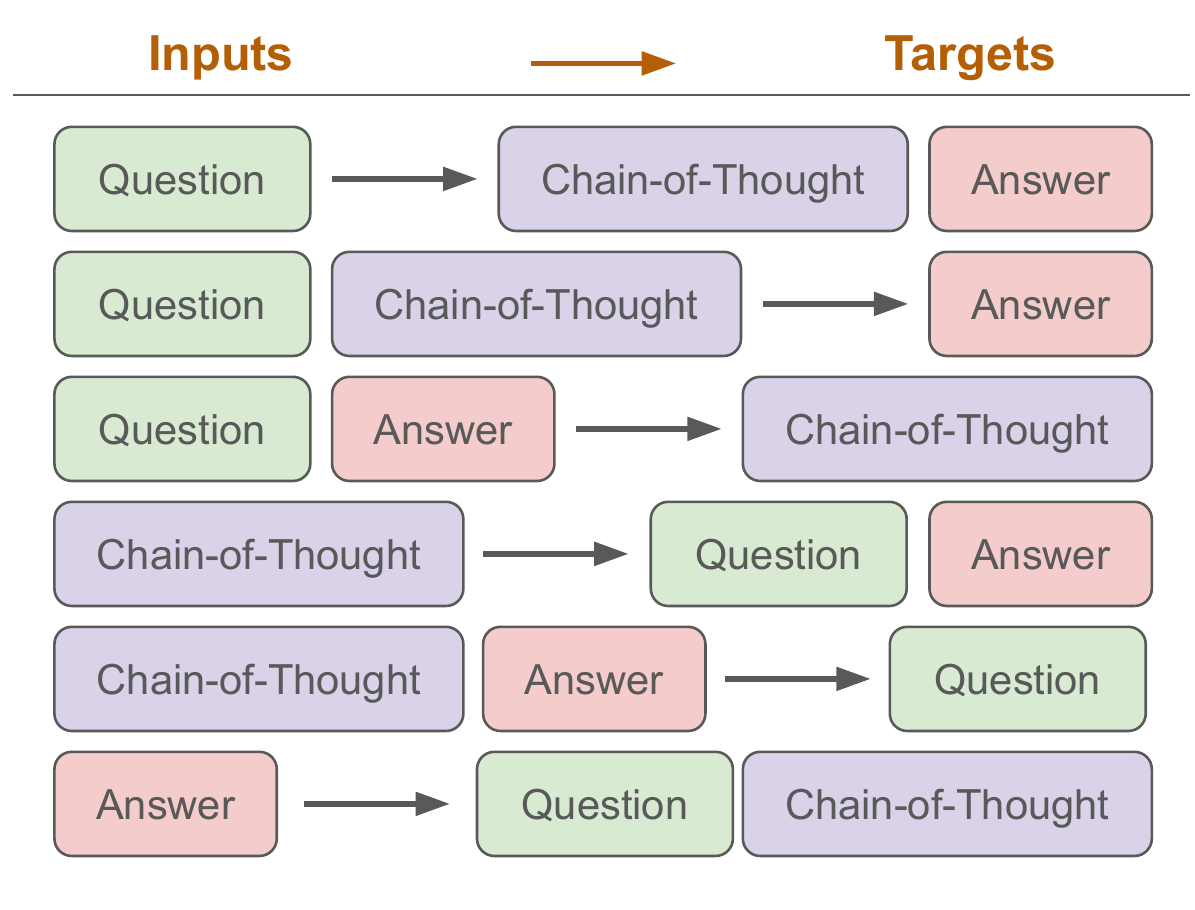}
    \caption{
    \small
    \textbf{Input Inversions permutations for a Zero-Shot Chain-of-Thought example.} Each is accompanied by a corresponding instruction template that prompts the model with what the input is, and what to predict as the targets.
    }
    \vspace{-3mm}
    \label{fig:cot-ii}
\end{figure}

For the input inversion experiments we note that Flan 2021, P3++, and Super-Natural Instructions already implicitly include tasks that have been inverted, e.g. question answering to question or context generation.
Consequently, we choose to also create input inversions for the remaining datasets in the \flantwo{} collection, including for the Dialog, Program Synthesis, and Chain-of-Thought tasks.

As examples: for Dialog tasks, we write template instructions asking for the previous conversational history from the current dialog turn; for program synthesis we ask for the coding question which the code solves; and for Chain-of-Thought we include every permutation of the query-answer-explanation triple, where at least one of the three appears as the in output.
An illustration of Chain-of-Thought input inversion permutations are shown in \cref{fig:cot-ii}.

These inversions are mixed in with the existing tasks at a rate of 30\%, meaning for a Dialog task, 3 inverted examples will be generated for every 10 regular examples.
We choose this rate for simplicity, approximately mirroring prior work, and leave the large space of exploration for future work.

\end{document}